\documentclass[letterpaper, 10 pt, conference]{ieeeconf}  

\IEEEoverridecommandlockouts                              

\let\labelindent\relax
\usepackage{dblfloatfix}
\usepackage[utf8]{inputenc} 
\usepackage[T1]{fontenc}    
\usepackage{hyperref}       
\usepackage{url}            
\usepackage{booktabs}       
\usepackage{amsfonts}       
\usepackage{nicefrac}       
\usepackage{microtype}      
\usepackage{amsmath,amsthm}
\usepackage{subcaption}
\usepackage{wrapfig}
\usepackage{color}
\usepackage[font=small,labelfont=bf]{caption}

\usepackage{array}
\newcolumntype{L}[1]{>{\raggedright\let\newline\\\arraybackslash\hspace{0pt}}m{#1}}
\newcolumntype{C}[1]{>{\centering\let\newline\\\arraybackslash\hspace{0pt}}m{#1}}
\newcolumntype{R}[1]{>{\raggedleft\let\newline\\\arraybackslash\hspace{0pt}}m{#1}}

\usepackage{blindtext}
\usepackage{tcolorbox}
\usepackage{graphicx}

\usepackage{graphics} 
\usepackage{epsfig} 
\usepackage{amssymb}  
\usepackage{wrapfig}
\usepackage{authblk}
\title{Domain Randomization for Active Pose Estimation}

\author{Xinyi Ren${^\dagger}$, Jianlan Luo${^\dagger}$, Eugen Solowjow${^\ddagger}$, Juan Aparicio Ojea${^\ddagger}$, \\
Abhishek Gupta${^\dagger}$, Aviv Tamar${^*}$, Pieter Abbeel${^\dagger}$ \\

$^\dagger$ UC Berkeley \\
$^\ddagger$ Siemens Corp \\
$^*$ Technion; work done while at UC Berkeley\\
}

\newtheorem*{hypothesis*}{Working Hypothesis}

\usepackage[inline]{enumitem}
\newenvironment{inline_enumerate}{
\begin{enumerate*}[label={(\arabic*)}]
}{\end{enumerate*}}

\begin{document}

\maketitle

\begin{abstract}
Accurate state estimation is a fundamental component of robotic control. In robotic manipulation tasks, as is our focus in this work, state estimation is essential for identifying the positions of objects in the scene, forming the basis of the manipulation plan. However, pose estimation typically requires expensive 3D cameras or additional instrumentation such as fiducial markers to perform accurately. Recently, Tobin et al.~introduced an approach to pose estimation based on domain randomization, where a neural network is trained to predict pose directly from a 2D image of the scene. The network is trained on computer generated images with a high variation in textures and lighting, thereby generalizing to real world images. In this work, we investigate how to improve the accuracy of domain randomization based pose estimation. Our main idea is that active perception -- moving the robot to get a better estimate of pose -- can be trained in simulation and transferred to real using domain randomization. In our approach, the robot trains in a domain-randomized simulation how to estimate pose from a \emph{sequence} of images.
We show that our approach can significantly improve the accuracy of standard pose estimation in several scenarios: when the robot holding an object moves, when reference objects are moved in the scene, or when the camera is moved around the object.
\end{abstract}
\section{Introduction}
In the past decades, robots have become dominant in industrial automation. A recent trend in manufacturing is the move toward small production volumes and high product variability \cite{lasi2014industry}, where reducing the manual engineering for automation becomes important. For automating many industrial tasks, such as picking, binning, or assembly, accurate pose estimation is essential. In this work, we focus on model based pose estimation from RGB cameras. This setting is relevant to many industrial applications, where a 3D model of the objects can easily be obtained, while it does not require expensive hardware such as high-precision depth cameras ~\cite{ye2011accurate,choi2012voting}
, nor making modifications to the object such as adding markers~\cite{olson2011apriltag}. Methods using markers often require significant human effort and have limited accuracy when the marker is far away or perpendicular to the image plane.

\begin{figure}[!t]
    \centering
    \includegraphics[width=\columnwidth]{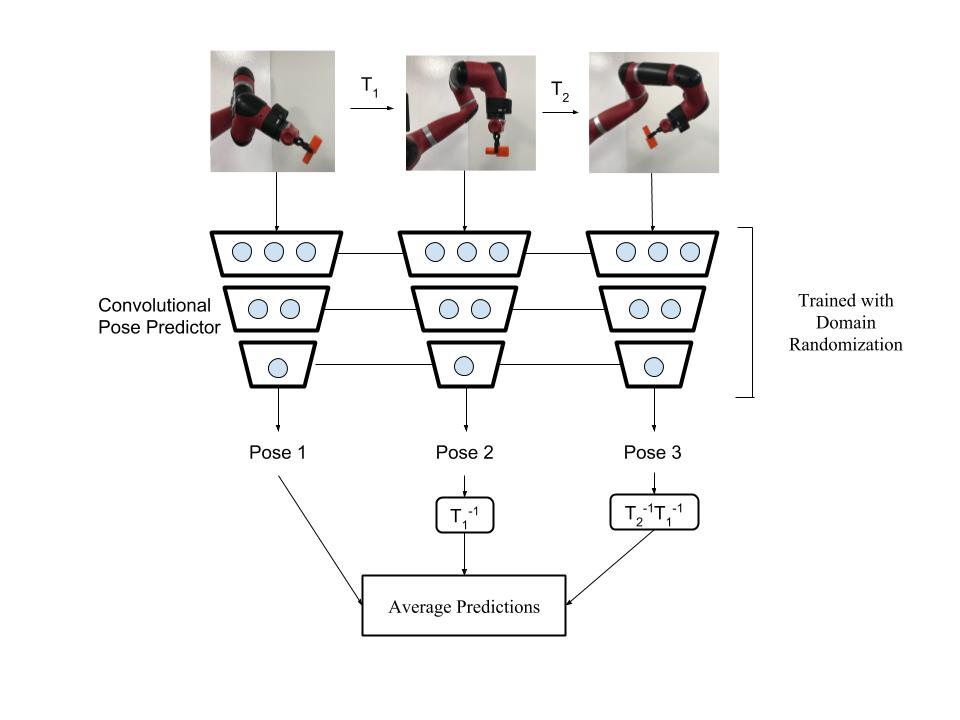}
    \caption{Inverse Transform Domain Randomization: We show that we can improve the accuracy of real world pose prediction with multiple images of a scene with known geometric transformations between object poses in the scenes.}
    \label{fig:teaser}
\end{figure}

While a number of methods have been proposed for model-based pose estimation using expensive depth cameras or extensive labelled datasets in the real world~\cite{xiang2017posecnn,tekin2018real}, the cost and manual effort required for these methods prevent them from being widely and easily applicable. Recently proposed methods proposing leveraging simulation as a tool for model-based pose estimation given accurate models of objects in an environment~\cite{tobin2017domain, sadeghi2016cad2rl, sim2realgoogle}. These methods are typically trained by leveraging known poses in simulation and training pose estimators which transfer effectively to the real world, bridging the simulation to reality gap.

In~\cite{tobin2017domain}, it was shown that domain randomization was able to reach a 1.5 cm error on 3D pose estimation. Many robotic tasks, such as assembly or bin placing, require a much higher precision. In this work, we investigate how to improve the accuracy of pose estimation based on domain randomization such that it is suitable for high precision robotic assembly tasks.

In this work, we aim to improve the accuracy of domain randomization based pose estimation by making the observation that robots don't have to be passive observers of a scene and can in fact interact with objects in the scene. We can perform a \emph{known geometrical transformation} to the scene, such as moving objects in the scene, moving the arm or distractors, or changing the camera angle. Since this transformation between scenes is known and applied by the robot, all of these data-points can be used in order to imporve the accuracy of pose predictions. We use this idea to propose a method for \emph{active} pose estimation, which exploits the fact that being able to see an object from different angles and in different positions leads to a more accurate and robust predictions. In this way, we find that actually leveraging consistency among multiple different images of a scene ensures a much more accurate pose estimation compared to standard ensemble methods such as domain randomization.

Using models for active pose estimation transferred from simulation, we are able to decrease the average error predicted on real camera image from 2 cm to under 0.5cm, which is sufficient to enable a variety of high precision robotic manipulation tasks which were otherwise very challenging with current methods.

\section{Related Work}
Our work has several connections to a number of prior works from various fields
\paragraph{2D model based pose estimation}
Model based pose estimation of rigid objects from a 2D image has been studied extensively, largely building on a predefined feature points~\cite{dementhon1995model,skrypnyk2004scene,Srinivasa2011IJRR,inria2006}, edge detectors~\cite{harris1992tracking,drummond2002real}, or image templates~\cite{jurie2002hyperplane}. We refer to~\cite{lepetit2005monocular} for an extensive survey.
Most of these algorithms rely on careful selection of the features to track, or on textured surfaces for points matching, and a careful calibration of the RGB camera. Our approach is agnostic to these factors.

\paragraph{3D model based pose estimation} Using a depth camera, high precision pose estimation can be obtained~\cite{ye2011accurate,choi2012voting}. However, accurate depth cameras (e.\,g.\, a Photoneo) can be very expensive, limiting their use in many applications. Our approach only requires a 2D RGB image.

\paragraph{Fiducial markers} The use of fiducial markers has become popular in augmented reality and robotics applications~\cite{kato1999marker,olson2011apriltag,bergamasco2011rune}. However, in realistic industrial applications, adding  fiducials to objects may be undesirable, and the accuracy of fiducial based pose detection is limited for certain poses (for example, when the fiducial is perpendicular to the image plane). Our approach does not require any external modification of the object for pose detection.

\paragraph{Pose estimation based on supervised learning} Several recent studies learned to map an image directly to pose using deep convolutional neural  networks (CNNs)~\cite{xiang2017posecnn,tekin2018real}. While the CNN structure in these works is similar to ours, these works require a labeled training set for learning, which can be difficult to obtain. The domain randomization approach, in contrast, generates its own training data by rendering in simulation.

\paragraph{Active perception}
The study of active perception~\cite{bajcsy1988active,bajcsy2018revisiting} concerns how a robot should take actions to better estimate parameters of its environment. To our knowledge, our work is the first to study active perception in a simulation-to-real setting.

\paragraph{Domain randomization} The gap between simulation and reality has been challenging robotics for decades. Recent work on trying to bridge this gap learns a decision making policy in simulation that works well under a wide variation in the simulation parameters, with the hope of learning a robust policy that transfers well to the real world. This idea has been explored for navigation~\cite{sadeghi2016cad2rl} and pose estimation~\cite{tobin2017domain}, by varying visual properties in the scene, and also for locomotion~\cite{mordatch2015ensemble} and grasping~\cite{tobin2017grasping}, by varying dynamics in simulation. In this work, we consider variation in the visual domain, and combine domain randomization with active perception, to improve its accuracy in pose detection.



\section{Problem Formulation and Preliminaries}
We consider a model-based rigid body pose estimation problem. In our setting, we assume that we have geometrical 3D models of an object $x$ and some reference object $y$.
Let $O_y$ denote a coordinate frame relative to $y$, and let $P_x$ denote the 6D pose of $x$ in the coordinate frame $O_y$. We are given an image of the scene $I$, that contains $x$ and $y$, and our goal is to estimate $P_x$ from the image.

\subsection{Pose estimation based on Domain Randomization}\label{ssec:pose_DR_background}
Tobin et al.~\cite{tobin2017domain} proposed a domain randomization method for solving the pose estimation problem described above. In this method, a 3D rendering software is used to render scene images with different poses of $x$ and $y$, and random textures, lighting conditions, camera orientations, and camera parameters. Let $D = \{I^1,P_x^1...,I_N,P_x^N\}$ denote the data set of the rendered images and matching object poses (which are known, by construction). Supervised learning is then used to train a deep neural network mapping $I$ to $P_x$. Since the network is trained to work on various texture, camera, and lighting conditions, it is expected that it also works on real world images since their statistics would roughly fall under the extremely wide distribution that was trained on. By making the training distribution extremely broad in terms of components such as texture, camera, and lighting, this method is able to ensure generalization to real world test environments by reducing the covariate shift. Indeed, the method in~\cite{tobin2017domain} reportedly obtained an average 1.5\,cm error in predicting 3D pose on real world test images.

\section{Method}

In this work, we propose an active perception approach based on domain randomization. To motivate our approach, we start by discussing the working hypothesis underlying domain randomization:

\begin{hypothesis*}[Domain Randomization]
There exist a set of features that can be extracted from all images in the data and are sufficient for predicting the image label (pose). These features can also be extracted from real images and are sufficient for predicting the real label.
\end{hypothesis*}

This working hypothesis means that if the training data is sufficiently randomized, and the neural network is expressive enough, then with enough data, the model has to discover the features which are common to all images, and base its prediction only on these features (otherwise it would suffer a higher training loss on spurious correlations that it picks up on). In that case, the network predictions are likely to transfer well to the real world.

One may question whether such features should even exist. However, for pose prediction, we know that the relative pose $P_x$ is a purely geometrical property of the objects, and since we assume an accurate 3D model of $x$, then geometrical properties (e.g., relative sizes and shapes) should be maintained in all the rendered images and also in the real images. Thus, the network has the potential to learn predictions based solely on geometrical properties of objects, abstracting away any other visual cues such as textures and lighting, and such features should transfer well to real images.

As discussed thus far, this pose estimation process is done completely passively. The robot does not interact with objects in the scene, but simply observes a single image of the scene and needs to predict the pose. In this work, we provide a key insight that we \emph{can} in-fact interact with the scene, and apply known geometric transformations to objects in the scene. These transformations allow us to obtain a number of different images of the scene to estimate the pose of the object as the transformations are all \emph{applied by us}. In this sense, we propose an active procedure to improve pose estimation by interacting with the scene and using multiple images to make a better prediction.


\subsection{Active Perception based on Domain Randomization with Geometric Transformations}

Recall that in the standard domain randomization problem (Section \ref{ssec:pose_DR_background}), training data is in the form of image-pose pairs, $\{I,P_x\}$. Following the active perception paradigm~\cite{bajcsy1988active}, we can apply to the scene some \emph{known} geometric transformation, with the hope that it improves our perception capabilities. For example, consider a robotic arm grasping an object, and the problem of estimating the position of the object within the robot's gripper. In this case, we can move the gripper closer to the camera to obtain a better pose estimate.
Since we know the transformation applied when moving the gripper, we can potentially combine several images to obtain a better prediction. As another example, consider moving the camera to obtain a better view of the object.

Concretely, we define the \textbf{Domain Randomization with Geometric Transformations} problem (DR-GT). let $T_1,\dots,T_k$ denote a set of $k$ transformations that can actively be applied to the geometry of the scene \emph{both in the real world and in simulation}. In particular, we consider rigid body transformations applied to objects in the scene and to the camera~\cite{faugeras1993three}.
We propose to generate training data in the form of tuples $\{I, T_1(I),\dots,T_k(I),P_x, T_1(P_x), T_k(P_x)\}$, where, slightly abusing notation, we denote by $T_i(I)$ and $T_i(P_x)$ the rendered image and pose when applying transformation $T_i$ to the scene.
The supervised learning problem we consider now is learning a mapping from $I, T_1(I),\dots,T_k(I), T_1,\dots,T_k$ to $P_x$.

\begin{figure*}[t]
\textbf{~~~~~~~~~~~~~~~~~~~~~~~~~~~~~~~~~In Simulation ~~~~~~~~~~~~~~~~~~~~~~~~~~~~~~~~~~~~~~~~~~~~~~~~~~~~~~In Real Life}\par\medskip
\begin{center}
\includegraphics[width=2.5cm]{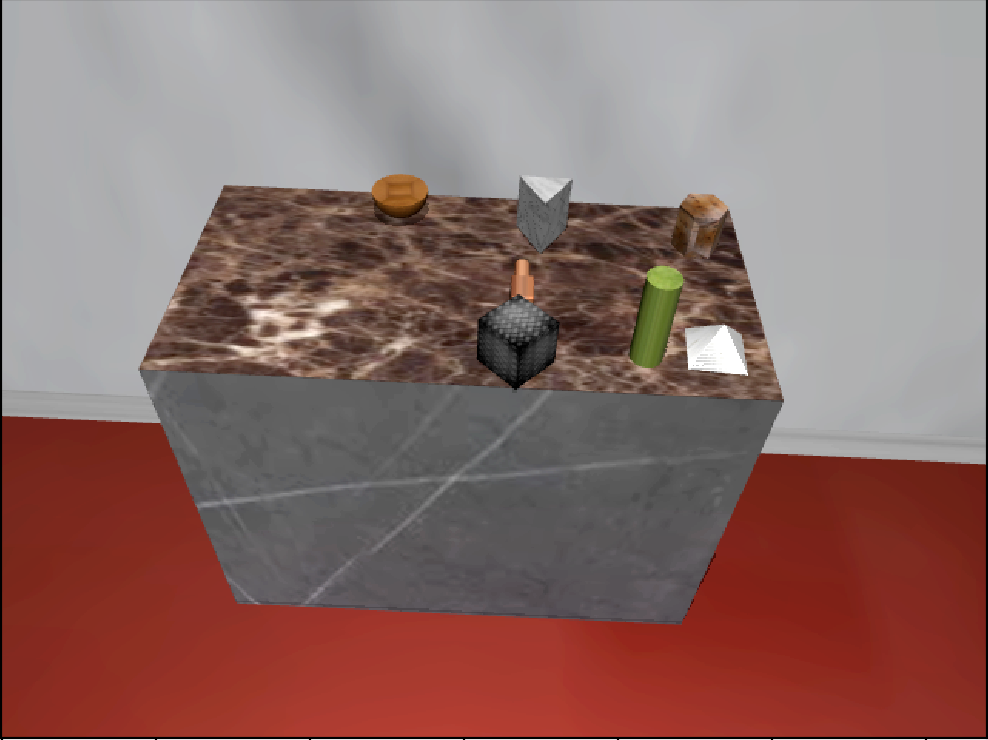}
\includegraphics[width=2.5cm]{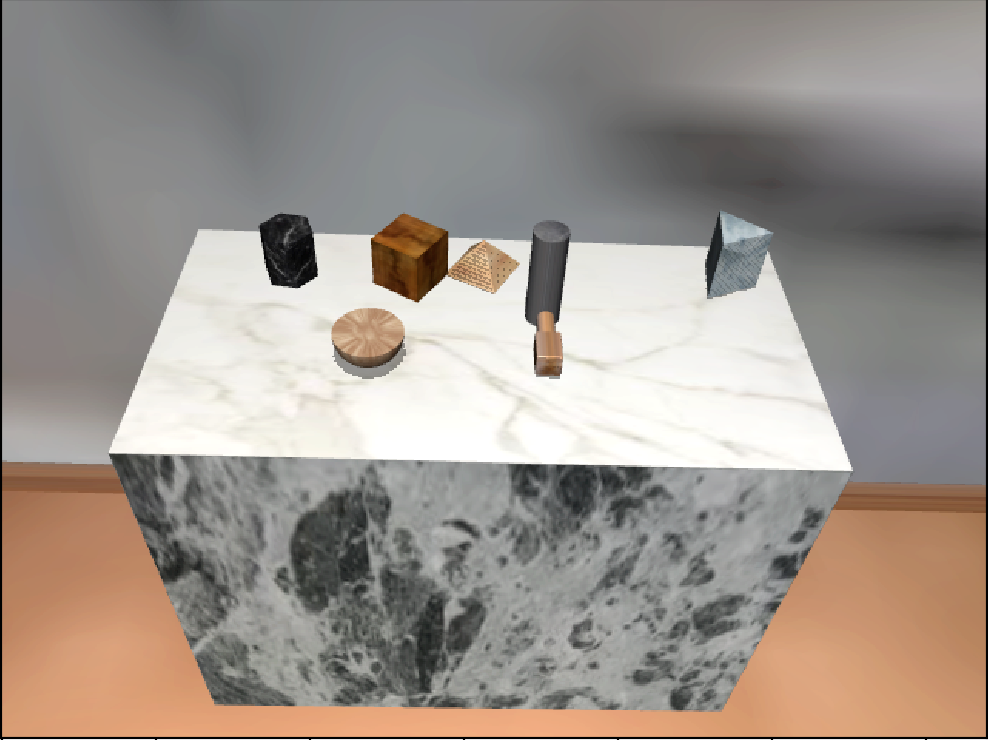}
\includegraphics[width=2.5cm]{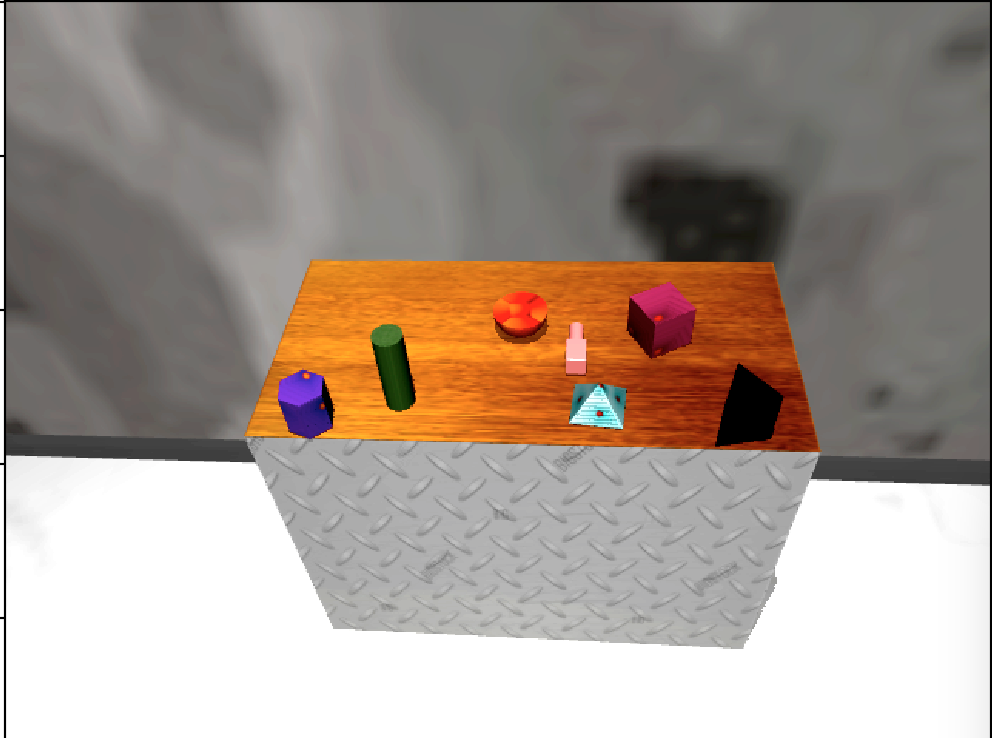}
\includegraphics[width=2.5cm]{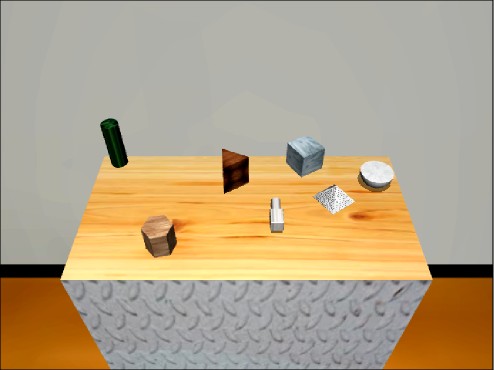}
\hspace{2cm}
\includegraphics[width=2.5cm]{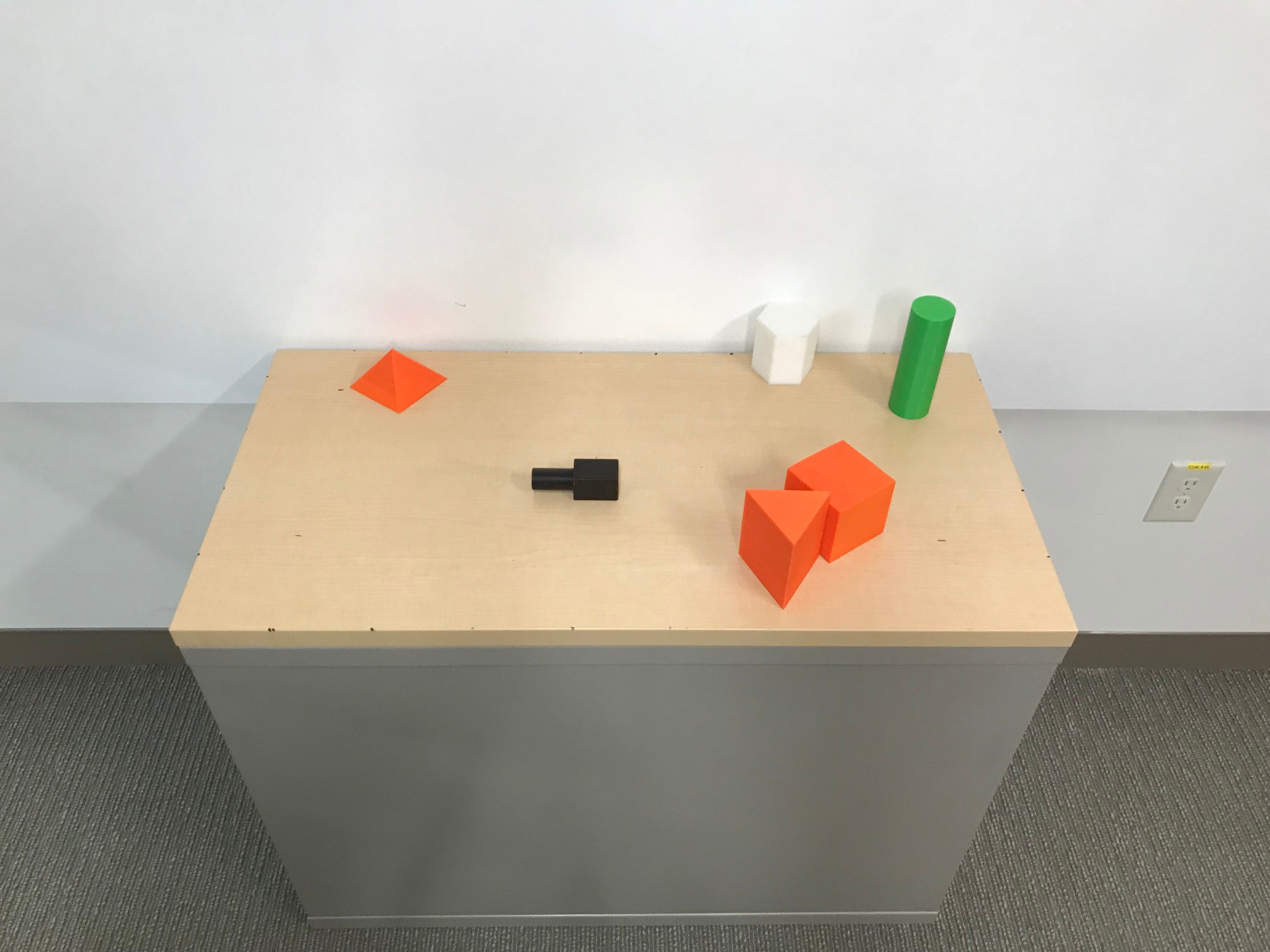}
\includegraphics[width=2.5cm]{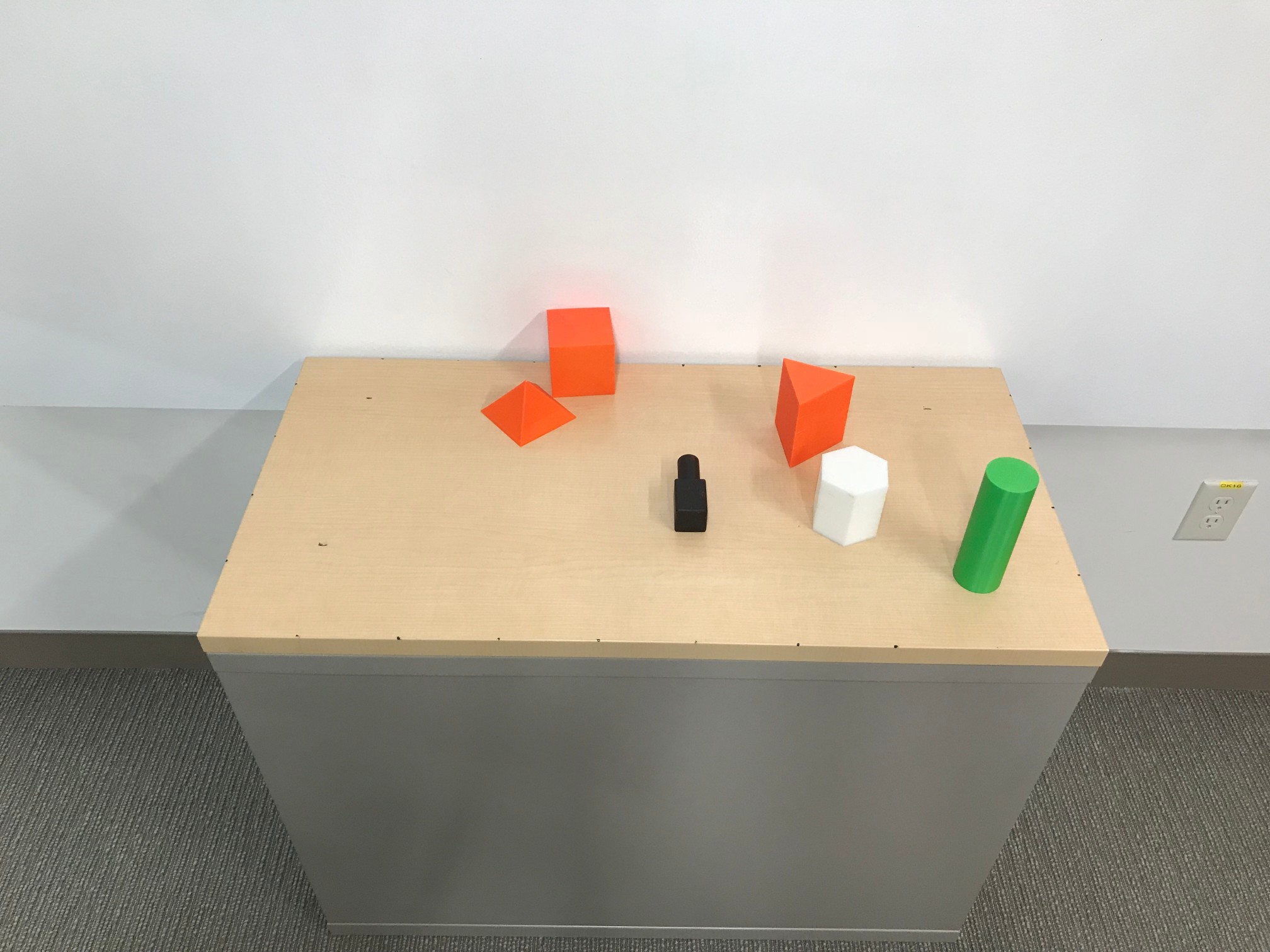}
\caption{Simulation to reality transfer with active reference object tabletop movement}
\end{center}
\end{figure*}

\subsection{Inverse Transform based Domain Randomization}
To solve the DR-GT problem, we propose the following method, based on inverse transforms. Let $T_i^{-1}$ denote the inverse transform of $T_i$.\footnote{We restrict our approach to transformations with a well-defined inverse, such as rotations and translations.} Let $f$ be the standard domain randomization mapping from $I$ to $P_x$. Then, we propose to calculate
\begin{equation}\label{eq:inverse_transform}
    \begin{split}
        P_{x;0} &= f(I),\\
        P_{x;1} &= T_1^{-1}(f(T_1(I))),\\
        &\dots,\\
        P_{x;k} &= T_k^{-1}(f(T_k(I))).
    \end{split}
\end{equation}
Note that for each $i\in 0,\dots,k$, the inverse transformation in \eqref{eq:inverse_transform} means that the prediction $P_{x;i}$ is an estimate of $P_x$. Therefore, we can predict $P_x$ as the sample average:
\begin{equation*}
    \hat{P_x} = \frac{1}{k+1}\sum_{i=0}^{k} P_{x;i}.
\end{equation*}
We term this method Inverse Transform based Domain Randomization (ITDR). We expect that as we enlarge the number of transformations $k$, the precision of ITDR improves.
While it is seemingly naive to use the sample average as a prediction, we found that it is surprisingly effective compared to more complicated methods with models which consider several images at once as input and produce a single pose estimate directly.

The key intuition behind using ITDR for improved estimation is that using known transformations in an environment allows us to use a wider data distribution to make several predictions of the same pose. Since several of these transformations provide easier to model prediction problems than the original problem, it makes the accuracy of the model significantly higher in the real world.

\section{Model Architecture}
In order to perform accurate pose estimation directly from images, we used a convolutional neural network architecture~\cite{goodfellow2016deep}. The neural network takes a single image as input, and generates a pose as output. In our experiments we investigated predicting 3DoF pose composed of 2DoF translation and 1DoF rotation. The model architecture takes in an RGB image through $16$ convolutional layers, each two convolutional layers is followed by a max-pooling operation and a ReLU nonlinearity. These convolutional layers are followed by $3$ fully connected layers with decreasing hidden units and ReLU nonlinearity. This architecture is similar to the one used in~\cite{tobin2017domain}, based on the VGG architecture~\cite{simonyan2014very} using  convolution layers pretrained on ImageNet.
The loss function for training this model is a combination of L1 regression loss for the 2DoF translation, and a cosine loss for the orientation, given by
$$L(x,\theta) = ||{x-\hat{x}}|| + ||{cos(\theta -\hat{\theta})-1}||,$$
where $x$ and $\hat{x}$ are the true and predicted pose, and $\theta$ and $\hat{\theta}$ are true and predicted orientation.

For active pose estimation, we pass a number of different images through the same network and then average the predictions \emph{after} applying a known rigid transform between them, as described in the ITDR algorithm.

\section{Experiments}
In this section we report our experiments on active perception using domain randomization. In our investigation, we aim to determine whether geometric transformations can give significantly better performance for model-based pose estimation in the real world. We designed our experiments to investigate whether using the robot to actively move elements of the scene yields gains in estimating the pose of objects in the scene. In particular, we investigate the performance of ITDR in the following situations:
\begin{inline_enumerate}
\item Moving reference objects in the environment,
\item Moving a robot manipulator holding an object, and
\item Moving a camera held by the robot.
\end{inline_enumerate}
We believe that these experiments showcase the potential of active perception used within domain randomization.

\subsection{Moving Reference Objects}
\label{sec:movingobjects}

We consider estimating the pose of a peg-shaped object $x$ resting on a table, similar to the experimental setting of~\cite{tobin2017domain} (see Figure \ref{fig:moving_reference} for more details). This is an important task for estimating where to grasp an object for further manipulation.
Since the table has a fixed height, and the object is at rest, the relevant degrees of freedom are 3-dimensional -- the 2D position of the object center and its orientation.\footnote{Note that Tobin et al.~\cite{tobin2017domain} only predicted the 2D position, while here we also predict orientation.}
We predict the pose of $x$ with respect to the green cylinder object $y$. For active perception, we consider moving the reference object $y$ between a fixed set of $4$ points in the corners of the table, and using ITDR to predict the pose from all images. We expect that actively moving objects in the scene will give us much more accurate pose estimation than if we had just a single image since the diversity of data to predict from is larger.

\begin{table}
\begin{center}
  \begin{tabular}{ | m{2.5cm}| m{0.8cm}| m{0.8cm}|m{1.2cm}|  }
    \hline
    Average error & x [cm] & y [cm] & $\theta$ [radians] \\ \hline
    Using only one image & 1.57 & 1.10 & 0.065\\ \hline
    Reference Object on Diagonal & 0.64 & 0.456 & 0.025 \\\hline
    Reference Object in Parallel & 1.17 &0.47  &0.038\\\hline
    Reference Object on Four Corners &0.62 &0.46 &0.037\\\hline
  \end{tabular}
\end{center}
\caption{Table showing the mean prediction error for pose prediction for the moving reference object scenario described in Section~\ref{sec:movingobjects}. This table shows that performance of pose prediction transferred form simulation to reality can be significantly improved by using the robot actively to modify the scene being considered. Prediction error is low when the reference object is placed at all four corners or on diagonal corners rather than along an edge of the table. This indicates that multiple images with known transformations do help reduce pose prediction error, and the choice of which transforms to use has a significant effect on the prediction error.}
 \label{fig:singleobject}
\end{table}

\begin{figure}[!t]
\begin{subfigure}[b]{0.48\columnwidth}
\includegraphics[width=\textwidth]{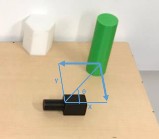}
\caption{Relative pose measurements with respect to green reference object (Section~\ref{sec:movingobjects}) }\label{fig:moving_reference}
\end{subfigure}~
\begin{subfigure}[b]{0.48\columnwidth}
\includegraphics[width=\textwidth]{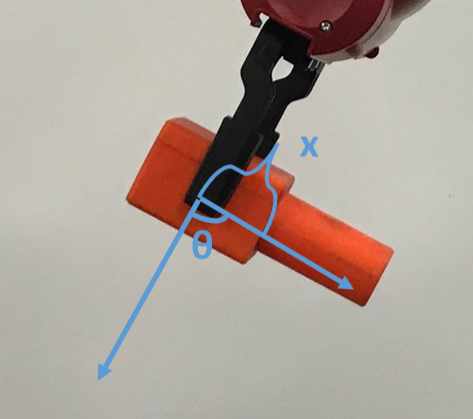}
\caption{Relative pose measurements when the object is grasped in the gripper (Section~\ref{sec:movinggripper})}\label{fig:moving_robot}
\end{subfigure}
\caption{Experiment setups: (a) estimate pose relative to movable reference object. (b) estimate pose relative to robot gripper.}\label{fig:moving_reference}
\end{figure}

\begin{figure*}[!t]
\textbf{~~~~~~~~~~~~~~~~~~~~~~~~~~~~~~~~~In Simulation ~~~~~~~~~~~~~~~~~~~~~~~~~~~~~~~~~~~~~~~~~~~~~~~~~~~~~~In Real Life}\par\medskip
\centering
\includegraphics[width=2.5cm]{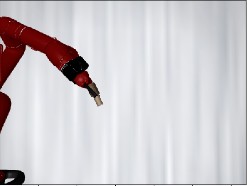}
\includegraphics[width=2.5cm]{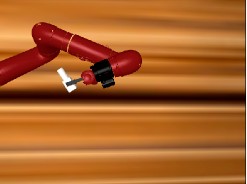}
\includegraphics[width=2.5cm]{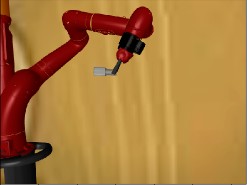}
\includegraphics[width=2.5cm]{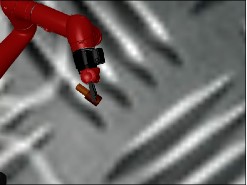}
\hspace{2cm}
\includegraphics[width =2.5cm]{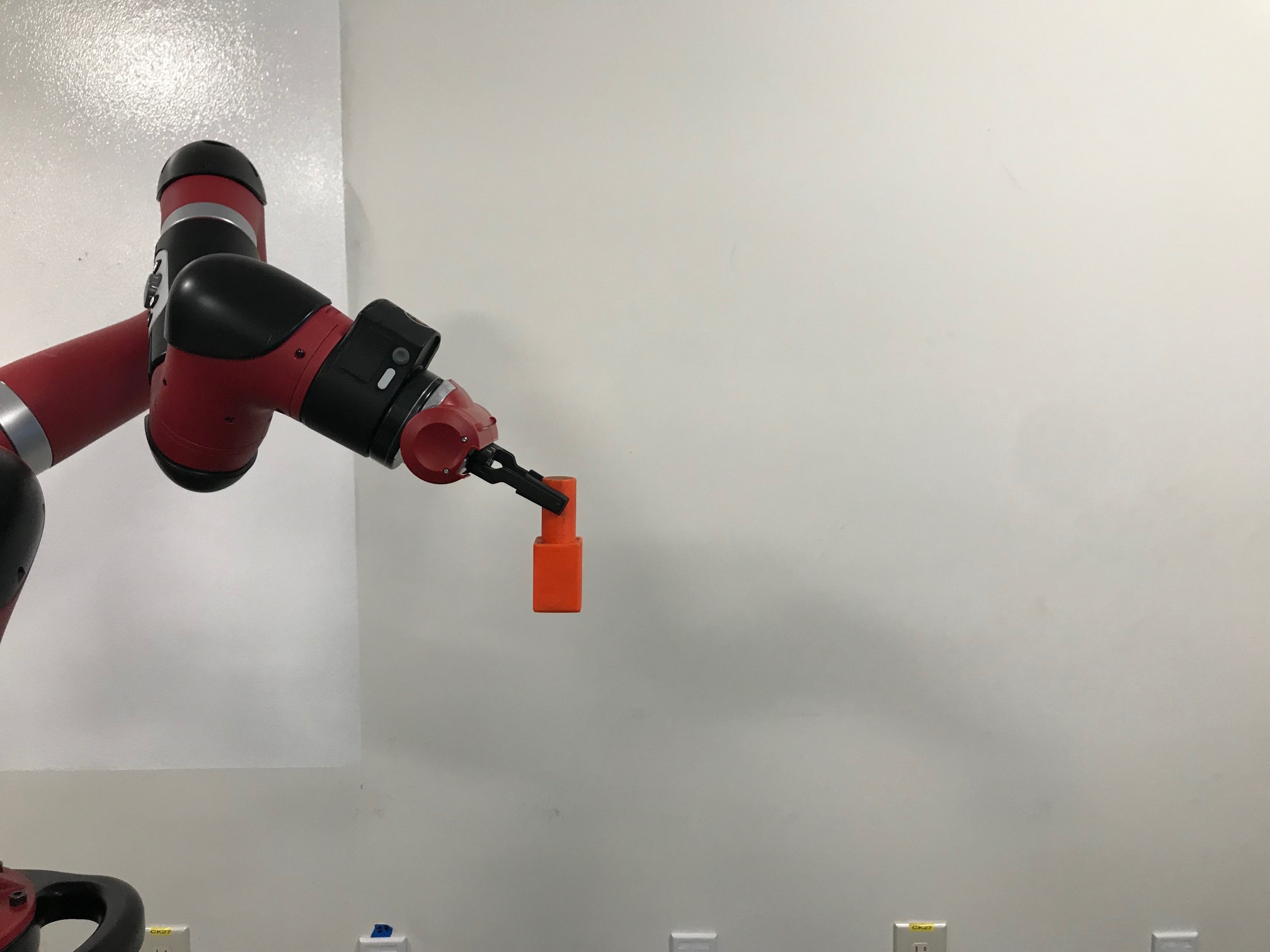}
\includegraphics[width =2.5cm]{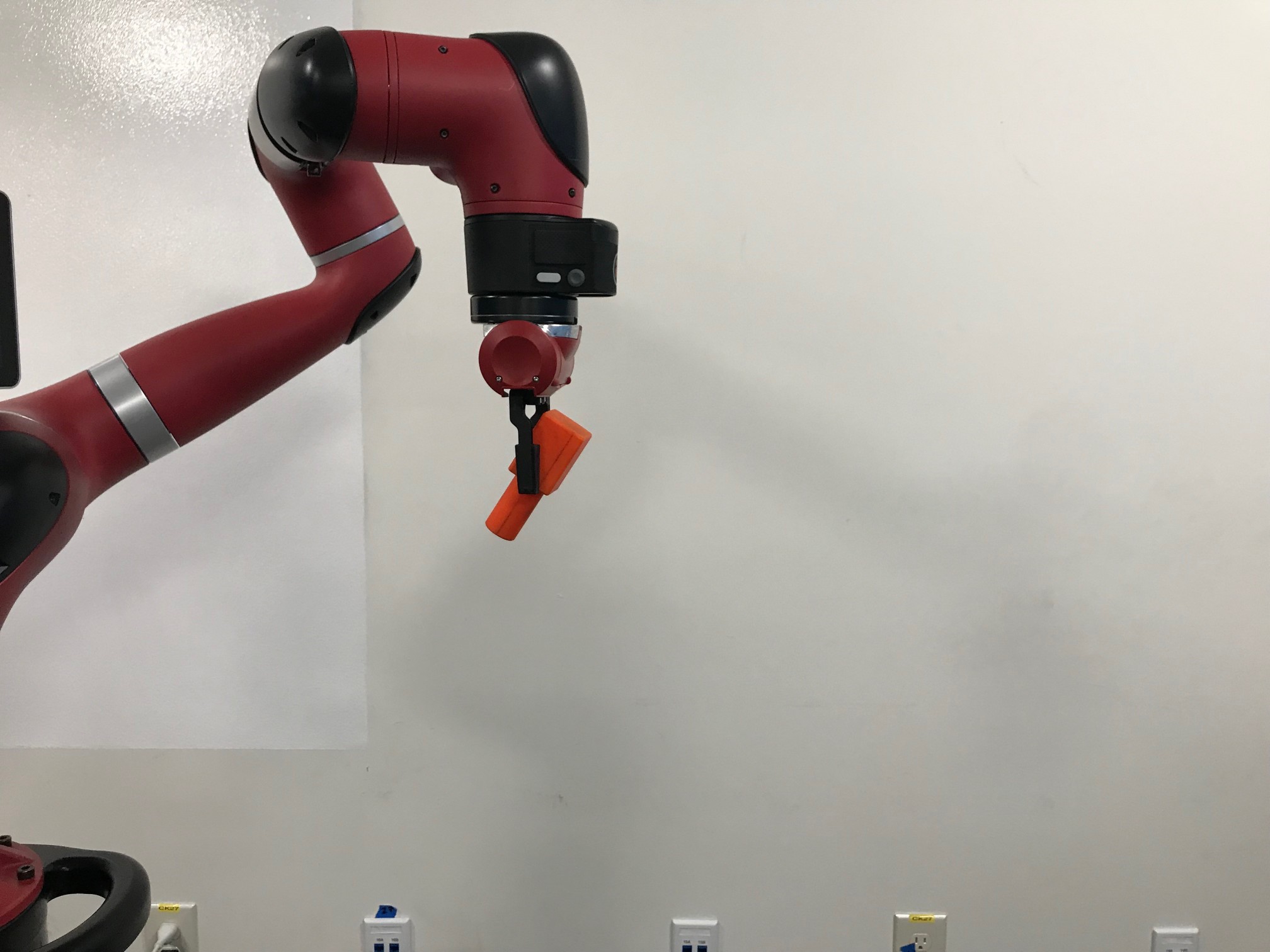}
\caption{Simulation to reality transfer with active gripper motion. Left: simulated images with domain randomization. Right: real images. Active perception here is based on moving the robot gripper. }
\end{figure*}

\begin{figure}[!h]
  \centering
  \begin{subfigure}[b]{0.45\linewidth}
    \includegraphics[width=\linewidth]{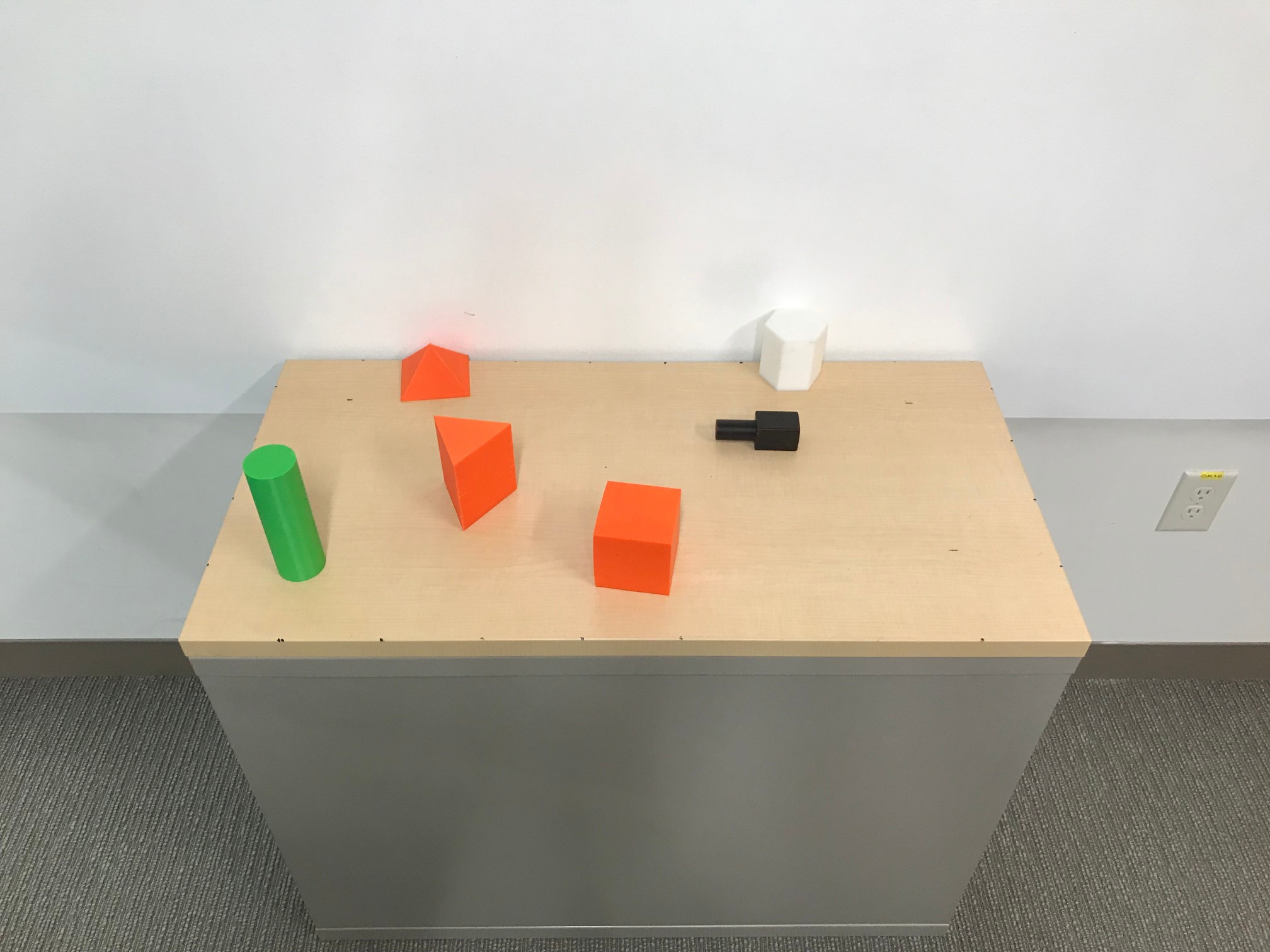}
  \end{subfigure}
  \begin{subfigure}[b]{0.45\linewidth}
    \includegraphics[width=\linewidth]{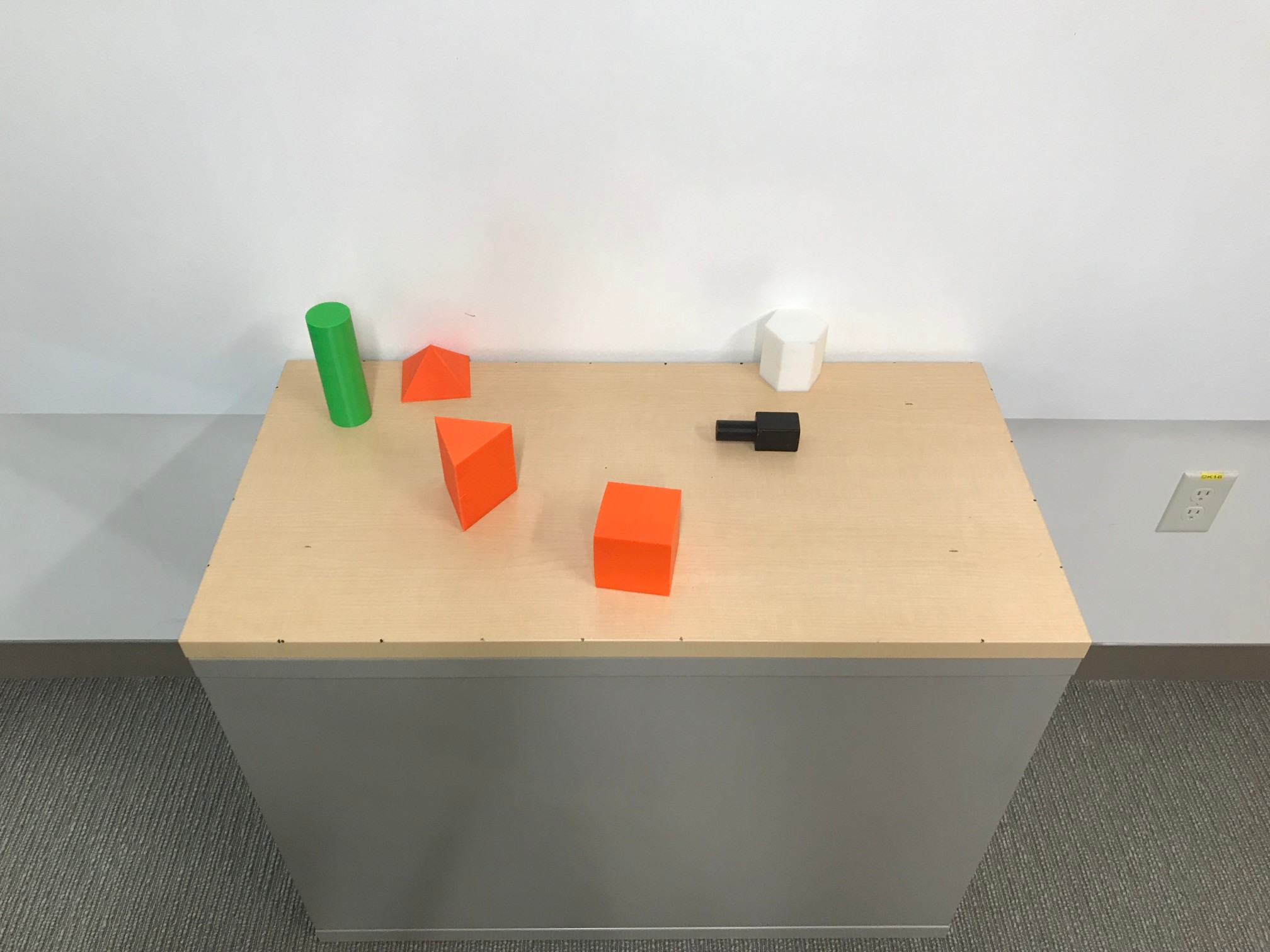}
  \end{subfigure}
  \begin{subfigure}[b]{0.45\linewidth}
    \includegraphics[width=\linewidth]{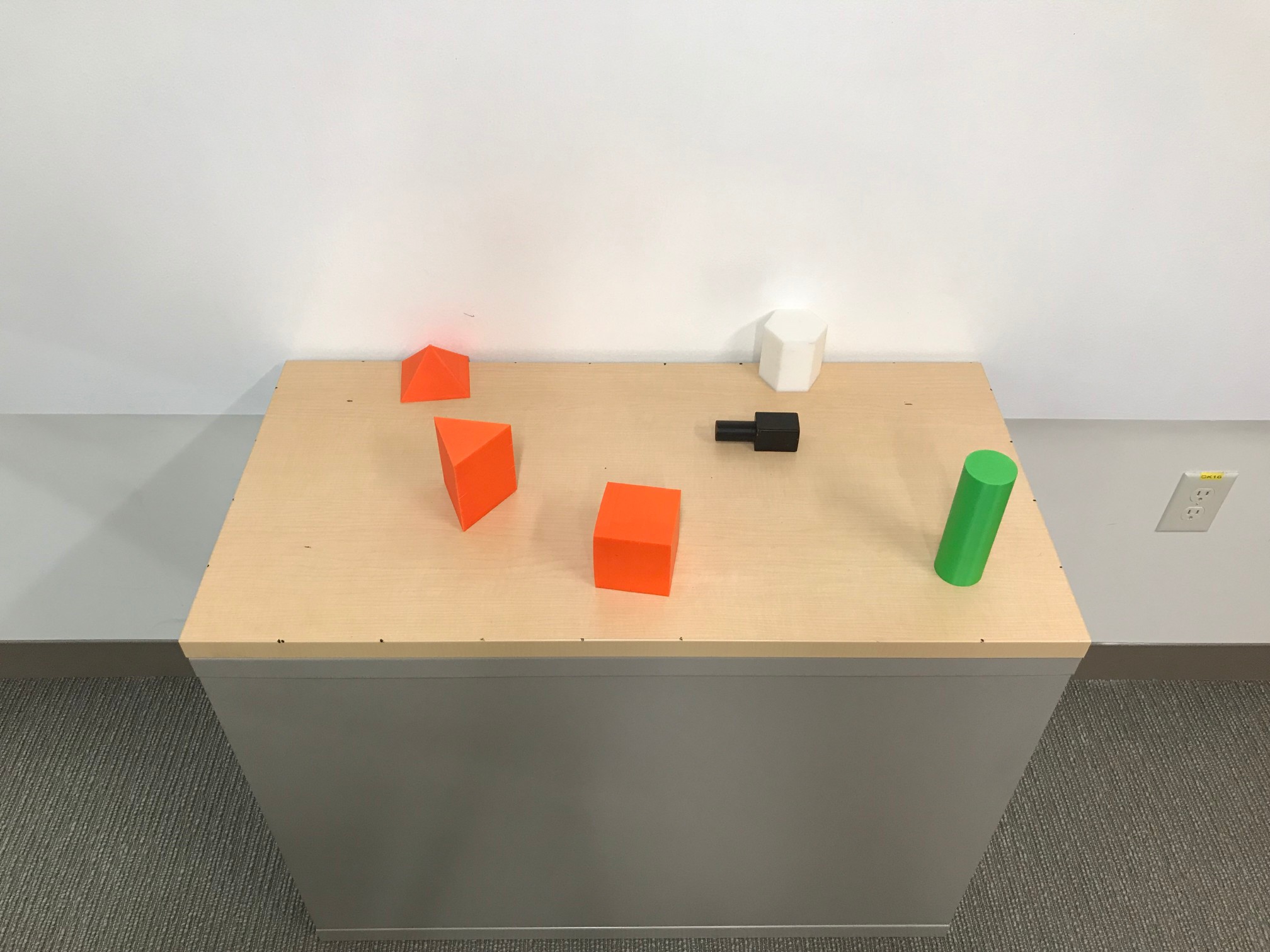}
  \end{subfigure}
  \begin{subfigure}[b]{0.45\linewidth}
    \includegraphics[width=\linewidth]{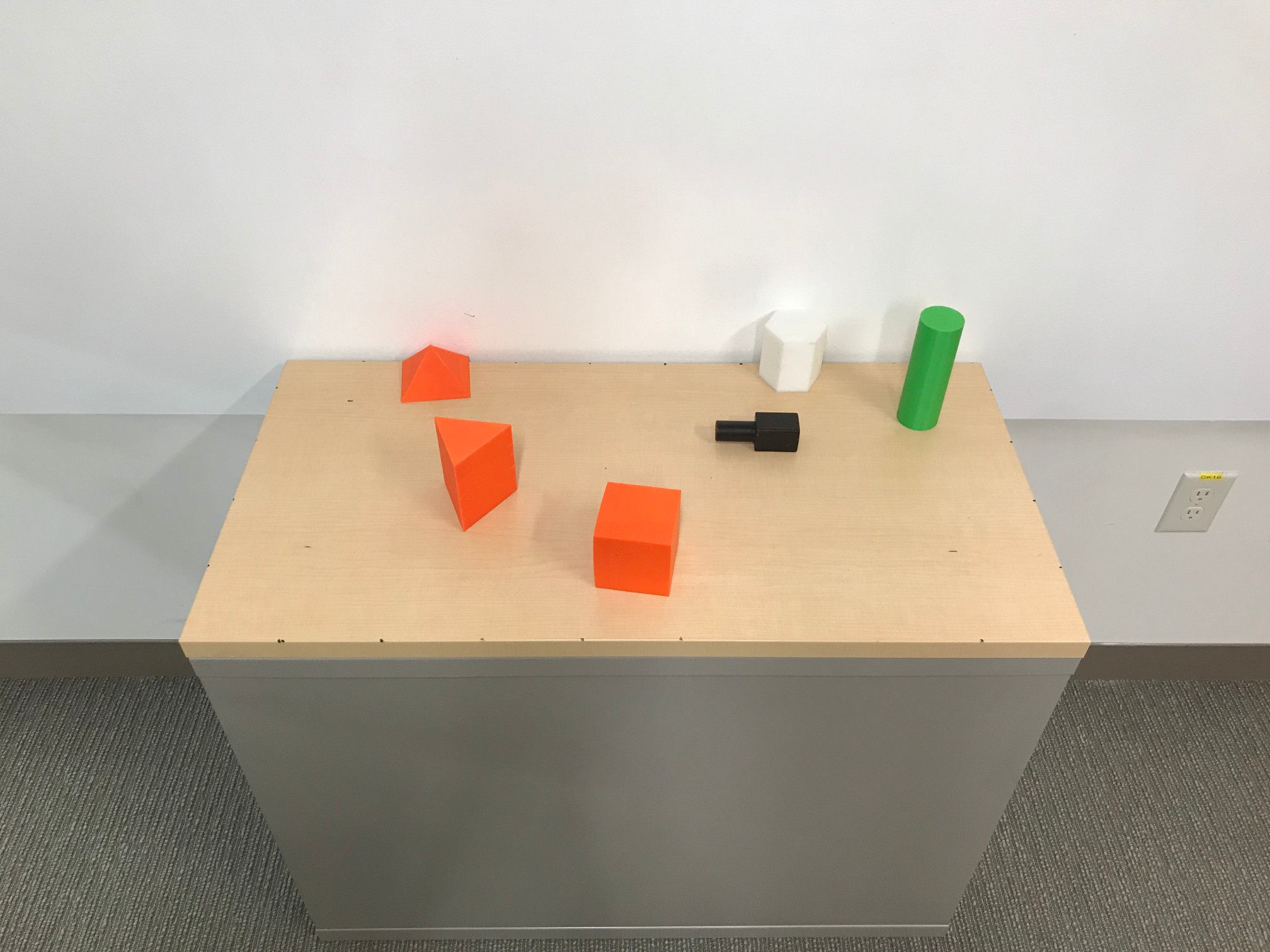}
  \end{subfigure}
  \caption{Different transform applied to the reference object as described in Section~\ref{sec:movingobjects}. The green cylinder is the reference object and we are estimating the pose of the black peg. As seen from these figures we can transform the position of the green cylinder to four different positions and use multiple images to improve pose estimation}
  \label{fig:cylindervariation}
\end{figure}

In Table~\ref{fig:singleobject} we show the results of pose estimation from a single image, and compare to using multiple images using ITDR. The improvement is on the order of 3x bringing down the estimation error to sub-centimeter ranges which enables a number of high precision applications.

We also investigate how to choose the best set of transformations for the reference object such that the pose estimation error is minimized. It is preferable for us to choose as few points as possible while ensuring accurate pose estimation. We choose pairs amongst the four points shown in Figure \ref{fig:cylindervariation} to evaluate whether particular transformations are more effective than others. In Table ~\ref{fig:singleobject} we report the error of each pair of possible reference object positions on real data. We see that the  real world error when moving object $y$ to the diagonal corners is significantly lower than if we moved the object to corners which share an edge. This is likely caused because it gives a more significant difference in the relative poses.

This experiment helps us understand the effect of moving objects in the scene actively in enabling better pose estimation. We see that simply moving reference objects in the scene and using the known geometric transformations allows us to estimate pose more accurately for real-world peg insertion tasks.

\subsection{Moving Robot Manipulator Holding an Object}
\label{sec:movinggripper}
From the results above, we see that moving reference objects in the scene to get a wider variety of relative poses significantly helps with pose estimation. Alternatively, we can consider a scenario where an object has already been grasped, but its position within the gripper is not known accurately. This would be a typical case when the pose estimation before grasping is not perfect. It is also important in robotic reinforcement learning experiments~\cite{levine2015learning}, where during learning, interaction with other objects in the environment can move an object that is grasped within the gripper.

\begin{figure}[!h]
  \centering
  \begin{subfigure}[b]{0.3\linewidth}
    \includegraphics[width=\linewidth]{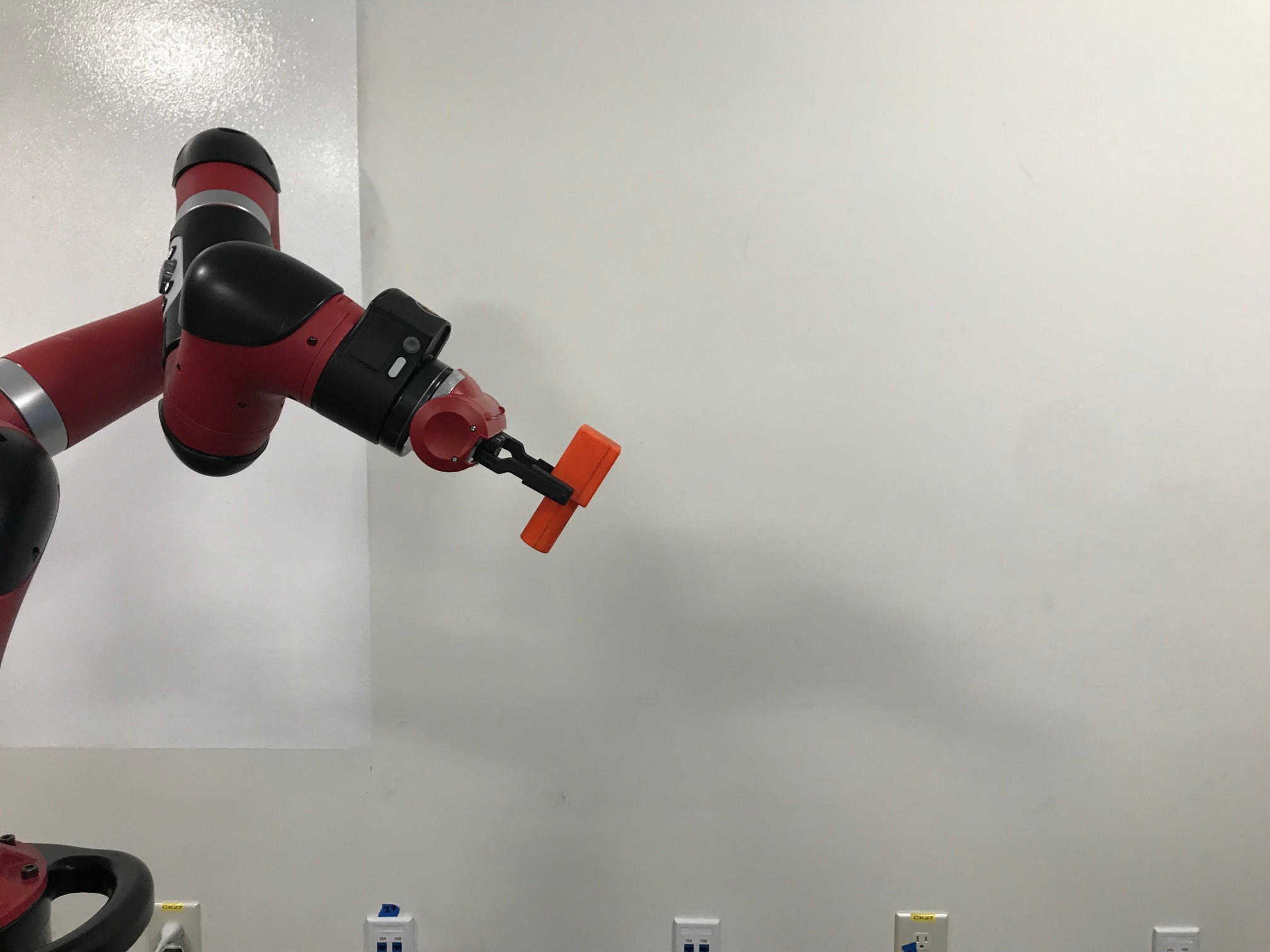}
  \end{subfigure}
  \begin{subfigure}[b]{0.3\linewidth}
    \includegraphics[width=\linewidth]{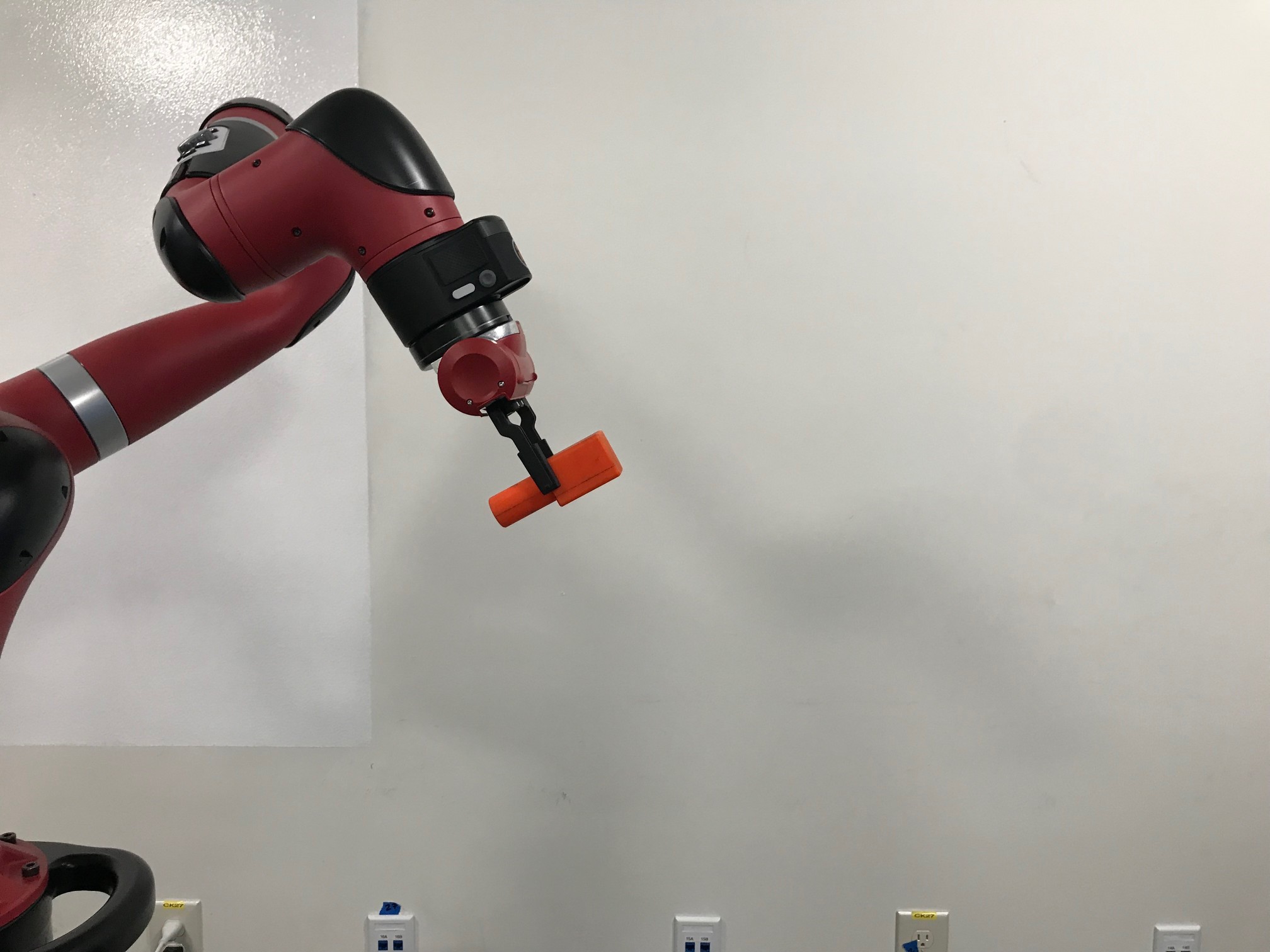}
  \end{subfigure}
  \begin{subfigure}[b]{0.3\linewidth}
    \includegraphics[width=\linewidth]{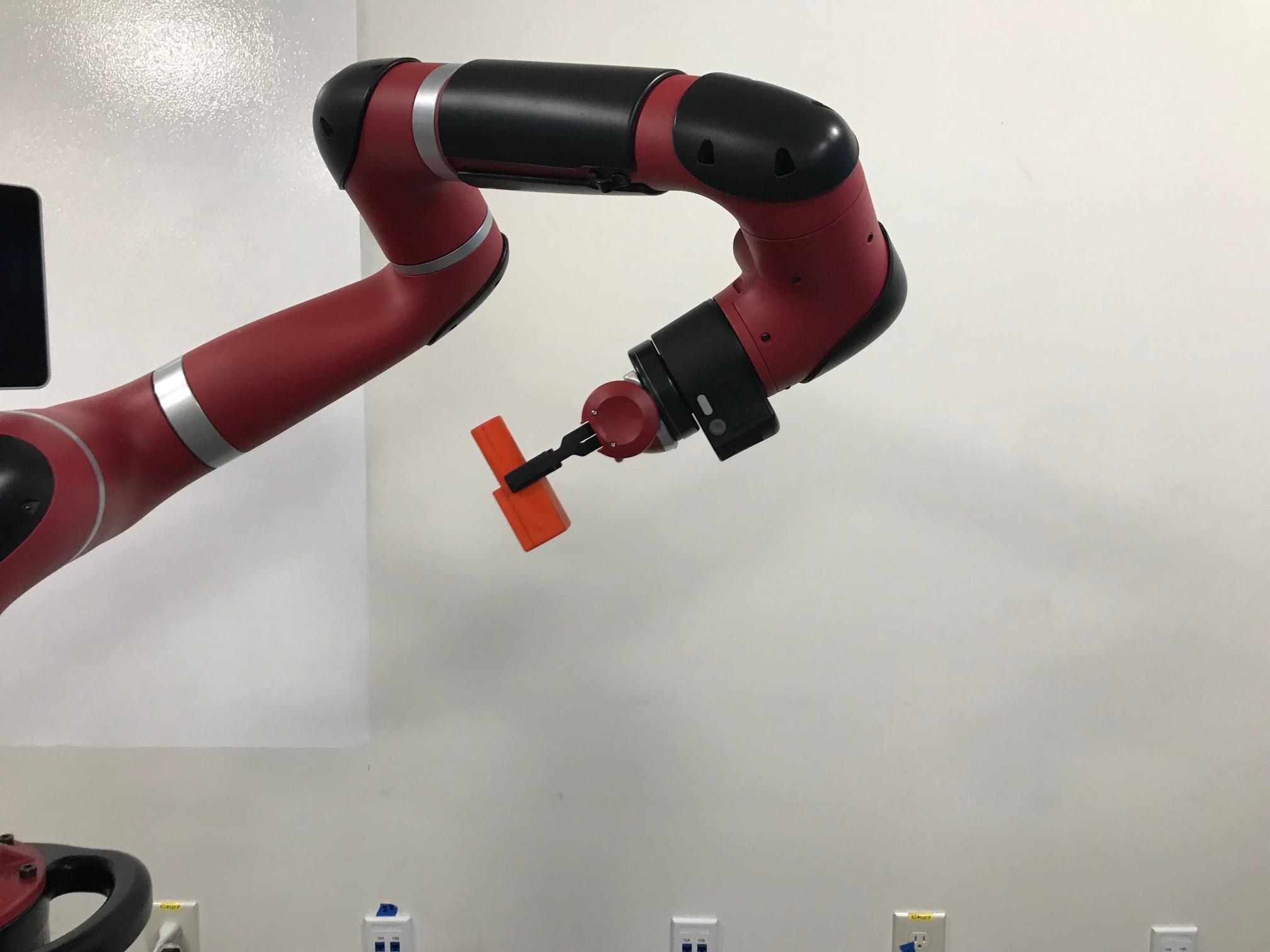}
  \end{subfigure}
    \caption{Different transform applied to the gripper with an object gripper in it. We want to estimate the exact relative position of the peg with respect to the gripper. As seen from these figures we can move the gripper to many different positions, with known transformations. We can use the additional viewpoints to improve the accuracy of pose estimation}
  \label{fig:grippervariation}
\end{figure}

As in the previous section, the object $x$ is peg-shaped, while the reference object $y$ in this case is the robot gripper. We estimate a 2-dimensional pose: the distance of the center of the object from the gripper, and its orientation within the gripper. The measured quantities are depicted in Fig \ref{fig:moving_robot}. These are challenging to estimate with extreme precision but are extremely important for the tasks we consider.

For active perception, in this case we move the robot gripper between a set of 5 fixed positions, and use ITDR to estimate the pose from all images. The different movements of the gripper show the camera different elements of the object itself, which is likely to help with better pose estimation since the model can latch on to different parts of the object. We find that this strategy indeed helps with pose estimation in the real world. We are able to identify the offset and the angle of the object grasped within the gripper significantly more accurately. ITDR performs significantly better than the baseline of simply using a single image and a model trained with domain randomization. As we can see from Table~\ref{fig:grippermove}, the x position error is improved by around $20\%$ and the angle accuracy is improved by around 3$\times$, from $0.129$ to $0.047$. Additionally, we find that using fewer gripper locations leads to worse performance. This suggests that using the multiple images does indeed improve performance and scales with using more images for estimation.

Note that in this setting, the object does not change pose with respect to the gripper, therefore the inverse transformations in ITDR are just the identity.

\begin{table}
\begin{center}
  \begin{tabular}{ | m{3.0cm}| m{0.8cm}| m{1.2cm}|  }
    \hline
    Average error & x [cm]  & $\theta$ [radians] \\ \hline
    one image & 0.30  & 0.129\\ \hline
    five images & 0.26  &0.047 \\\hline
    two images &0.27 &0.086\\\hline
  \end{tabular}
\end{center}

\caption{Table showing the mean prediction error for pose prediction for moving gripper scenario described in Section~\ref{sec:movinggripper}. We see that using multiple images with known geometric transformations is able to significantly reduce the angle error and provide some improvements in the estimation of the offset of the gripped object as well.}
\label{fig:grippermove}
\end{table}

\begin{figure}[!h]
\includegraphics[width=\linewidth]{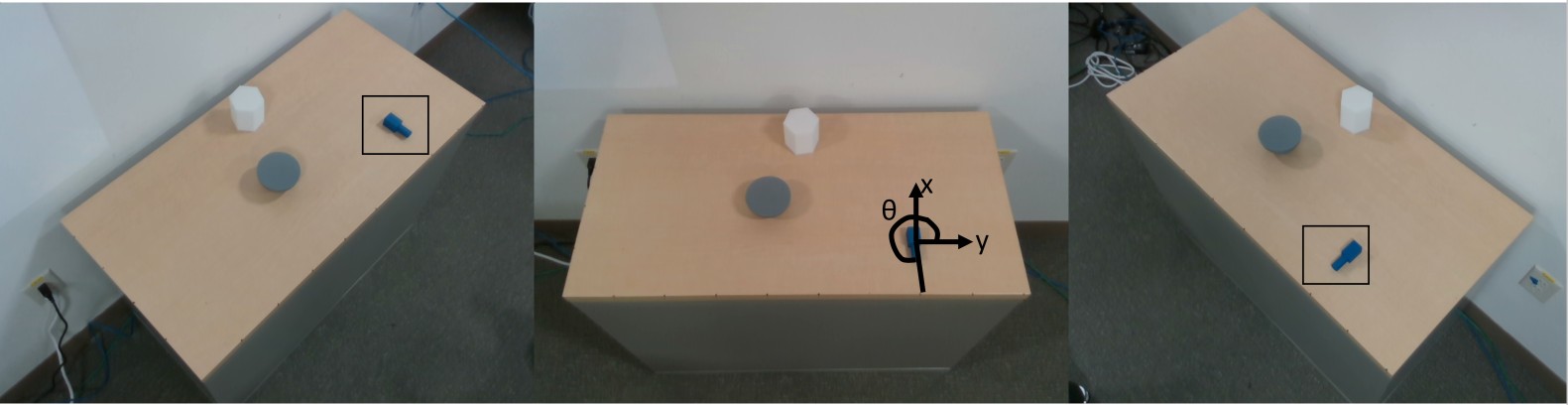}
\caption{Different camera angles of the same setting with target object boxed up. We want to estimate the position and orientation of the target object relative table corner. We can use multiple observing angles to explore the geometric properties of the target object and achieve a better pose estimation.}
\label{fig:exp3setup}

\end{figure}

\subsection{Moving a Robot-Held Camera}
\label{sec:movingcamera}
In this experiment, we demonstrate that actively moving the camera can improve the pose prediction performance. Figure ~\ref{fig:exp3setup} depicts our experimental setup.

As in Section \ref{sec:movingobjects}, we estimate the pose of a peg-shaped object $x$ resting on a table among various distractor objects, and a fixed reference object $y$. To obtain more contrasting result, the peg is $0.7^3$ times smaller than the one use in \ref{sec:movingobjects}. Here, we mounted our camera to the robot's end effector, and we actively change the viewpoint by moving the robot arm to various positions. In particular, we chose a set of three fixed positions in which we can view the object from. Note that similarly to Section \ref{sec:movinggripper}, the object does not change pose with respect to the reference, therefore the inverse transformations in ITDR are just the identity.

\begin{figure}[!h]
\includegraphics[width=\linewidth]{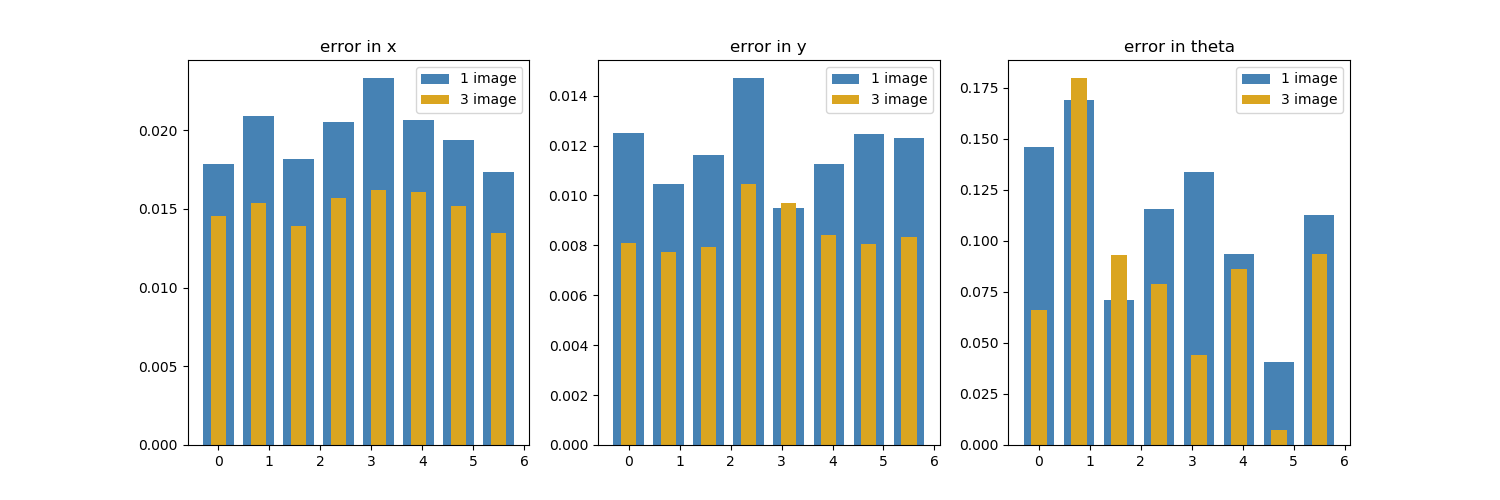}
\caption{Distribution of the error in horizontal direction, vertical direction, and $\theta$ with respect to the orientation of the object($\theta$). Using one image, we see a high prediction error  when the peg is orientated away from the camera ($\theta$ around 3.14) due to the asymmetric shape of the peg. When using three images, the prediction for this previous difficult object orientation is significantly improved.  }\label{fig:errortheta}
\end{figure}

In Table~\ref{fig:movecameraresult}, we show the results for our method. We observe a significant improvement in pose prediction for the x, y coordinates the object. To better understand these results, in Figure ~\ref{fig:errortheta} we plot the prediction error as a function of the orientation of the object. We observe that for some orientations, estimating the pose from a single viewpoint is very difficult, which is attributed to the asymmetric shape of the peg -- when placed such that it is perpendicular to the camera plane, most of the object is occluded, making it difficult to estimate a correct orientation. In test cases, adding additional viewpoints significantly improves the results.

\begin{table}
\begin{center}
  \begin{tabular}{ | m{2.5cm}| m{0.8cm}| m{0.8cm}|m{1.2cm}|  }
    \hline
    Average error & x [cm] & y [cm] & $\theta$ [radians] \\ \hline
    Using only one image & 1.97 & 0.12 & 0.11\\ \hline
    Using three images &1.50 &0.08 &0.08\\\hline
  \end{tabular}
\end{center}
\caption{Table showing the mean prediction error for pose prediction for the moving camera scenario described in Section~\ref{sec:movingcamera}. We see that using multiple images taken from different point of view is able to reduce both the coordinates and angle error.}
 \label{fig:movecameraresult}
\end{table}

\section{Discussion and Future Work}

In this work, we explored the use of active perception within the domain randomization paradigm. We have shown that active perception strategies which are able to interact with objects in the scene with known geometric transformations can significantly improve the performance of pose estimation compared to passive perception approaches. In particular, we have reduced the 1.5cm pose estimation error in domain randomization state-of-the-art to less than 0.6cm, which can lead to new robotic capabilities in downstream tasks such as tight fitting assembly problems.

In future work, we intend to explore additional methods for improving the performance of domain randomization based pose estimation, for example, in a semi-supervised setting where unlabeled images from the real domain are available to the learning algorithm.

\section*{Acknowledgements}
This work was supported in part by Siemens Corporation.

\bibliographystyle{IEEEtran}
\bibliography{references}

\end{document}